\documentclass[letterpaper]{article} %

\ifdefined\aaaianonymous
    \usepackage[submission]{aaai2026}  %
\else
    
    \usepackage{aaai2026}              %
    
\fi

\usepackage{times}  %
\usepackage{helvet}  %
\usepackage{courier}  %
\usepackage[hyphens]{url}  %
\usepackage{graphicx} %
\usepackage{natbib}  %
\usepackage{caption} %
\usepackage{algorithm}
\usepackage{algorithmic}
\usepackage{amsfonts}       %
\usepackage{amsmath}
\usepackage{multirow} 
\usepackage{framed}
\usepackage{booktabs}       %
\usepackage{makecell}
\usepackage{subcaption}
\usepackage{cuted}
\usepackage{xcolor}
\usepackage{bbding}

\usepackage{newfloat}
\usepackage{listings}
\DeclareCaptionStyle{ruled}{labelfont=normalfont,labelsep=colon,strut=off} %
\floatstyle{ruled}
\newfloat{listing}{tb}{lst}{}
\floatname{listing}{Listing}

\ifdefined\aaaianonymous
    \title{Understanding Dynamic Scenes in Egocentric 4D Point Clouds}
\else
    \title{Understanding Dynamic Scenes in Egocentric 4D Point Clouds}
\fi

\author {
    Junsheng Huang\textsuperscript{\rm 1},
    Shengyu Hao\textsuperscript{\rm 2},
    Bocheng Hu\textsuperscript{\rm 1},
    Hongwei Wang\textsuperscript{\rm 1,}\raisebox{-0.4ex}{\textsuperscript{\normalfont *}},
    Gaoang Wang\textsuperscript{\rm 1,}\thanks{Corresponding author.}
}
\affiliations {
    \textsuperscript{\rm 1}Zhejiang University, Hangzhou, China \\
    \textsuperscript{\rm 2}China Tower Corporation Limited, Hangzhou Science and Technology Innovation Center, Hangzhou, China \\
    junsheng.24@intl.zju.edu.cn,
    victorhsy7@gmail.com,
    thebrandonhu@gmail.com, \\
    hongweiwang@intl.zju.edu.cn,
    gaoangwang@intl.zju.edu.cn
}

\usepackage{bibentry}

\begin{document}
\maketitle
\ifdefined\aaaianonymous
\else
\fi

\stripsep-9mm
\begin{strip}
  \centering
    \includegraphics[width=0.95\textwidth]{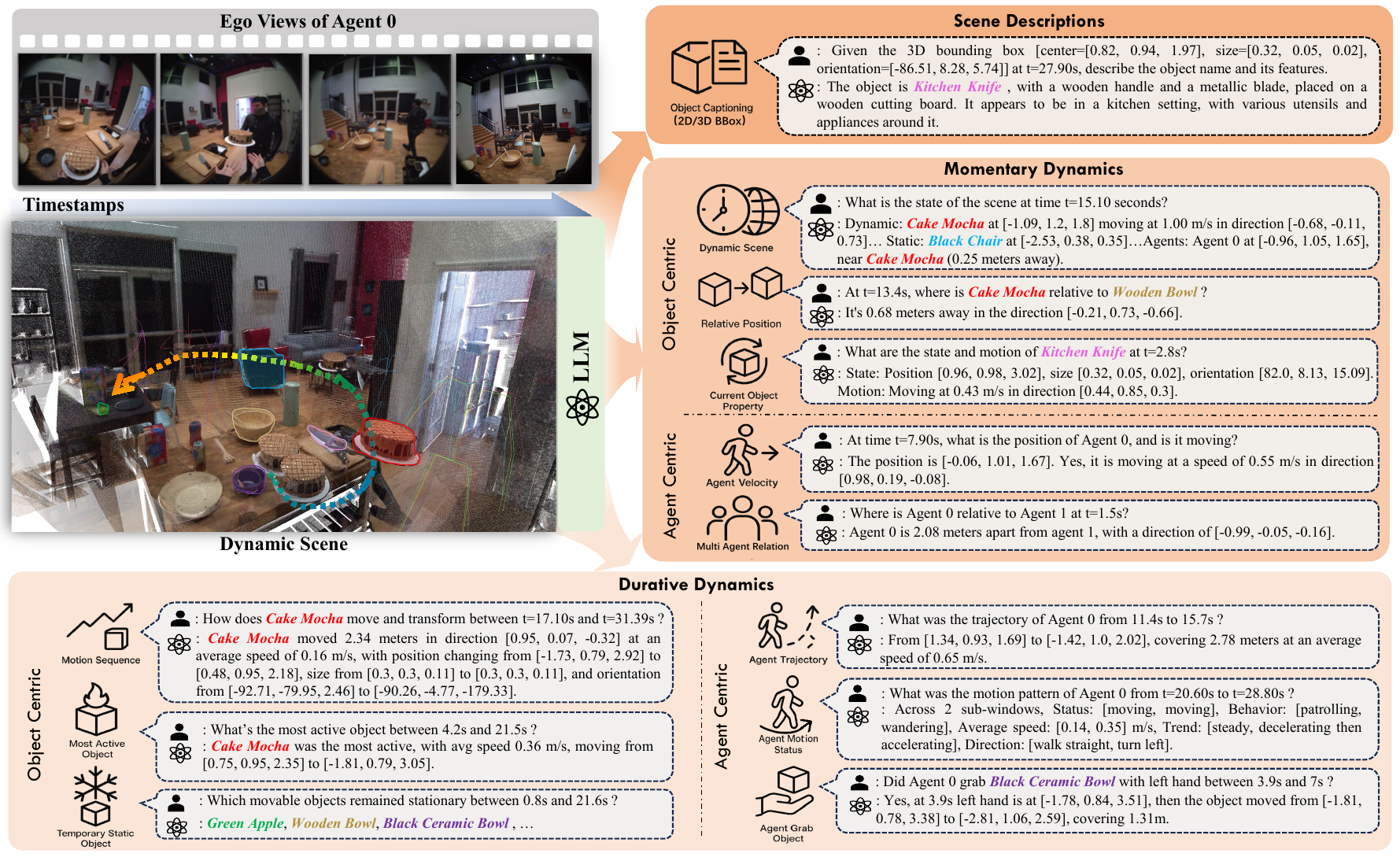}
    \captionof{figure}{We introduce a novel QA benchmark \textbf{EgoDynamic4D} and  an end-to-end spatio-temporal reasoning framework.}
    \label{fig:dataset-overview}
    \vspace*{16mm}
\end{strip}

\begin{abstract}

\begin{figure*}[ht]
    \centering
    \begin{subfigure}{0.26\linewidth}
        \includegraphics[width=\linewidth]{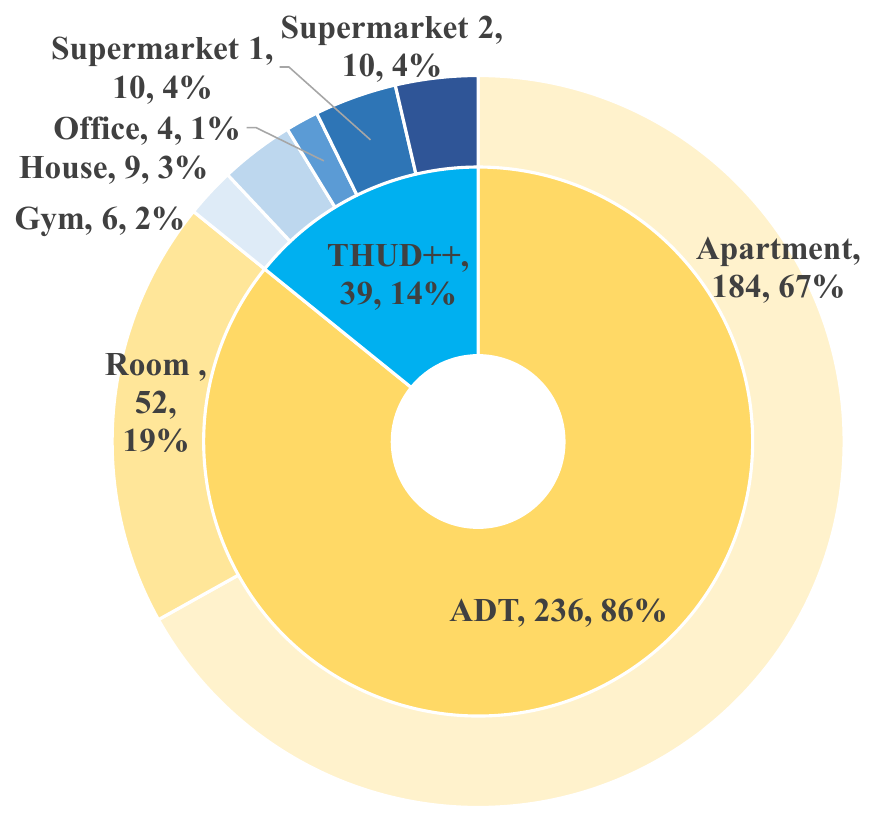}
        \caption{Data sources that provide videos captured in different scenes.}
        \label{fig:scene_distribution}
    \end{subfigure}
    \hfill
    \begin{subfigure}{0.34\linewidth}
        \includegraphics[width=\linewidth]{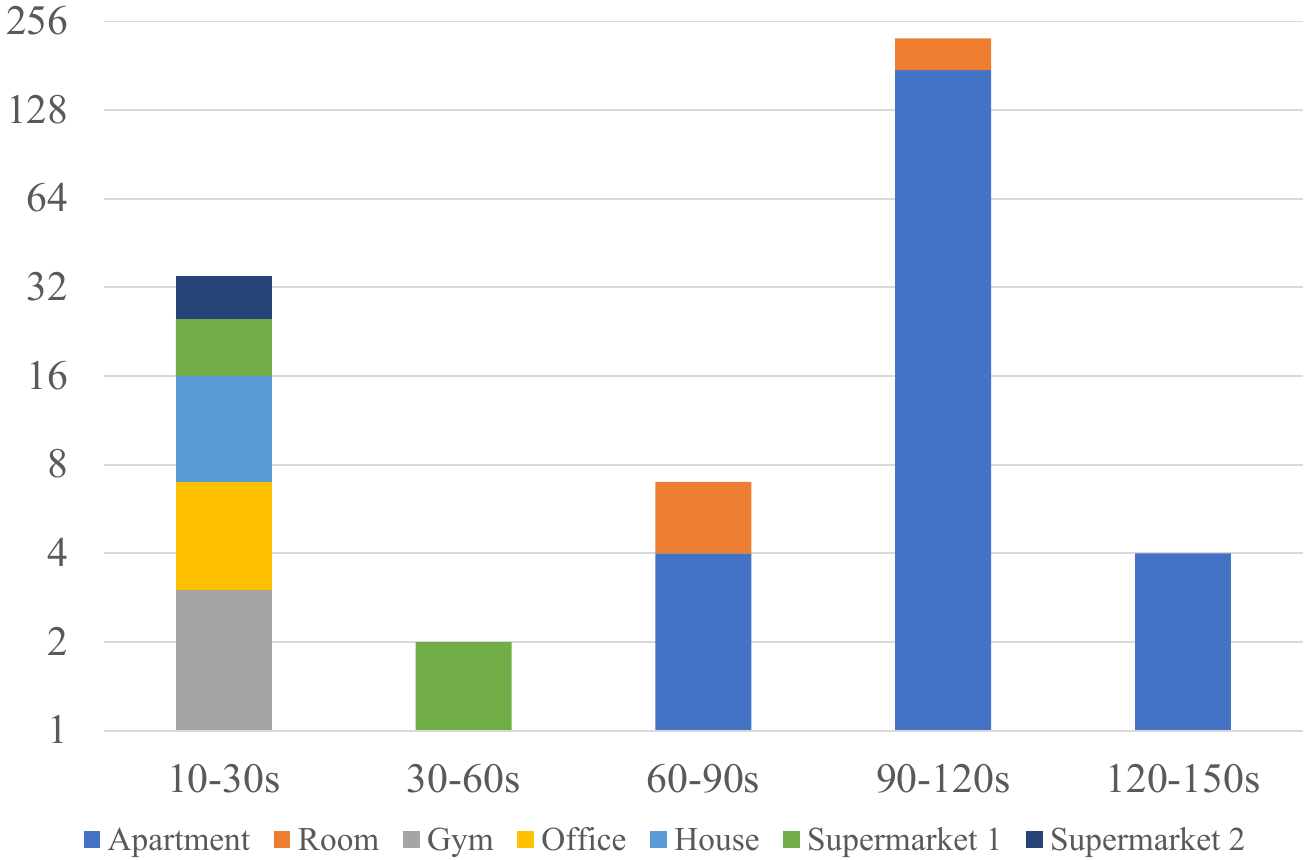}
        \caption{Duration of sequences. Peaks at 10-30s (38 sequences) and 90-120s (224 sequences).}
        \label{fig:temporal_distribution}
    \end{subfigure}
    \hfill
    \begin{subfigure}{0.30\linewidth}
        \includegraphics[width=\linewidth]{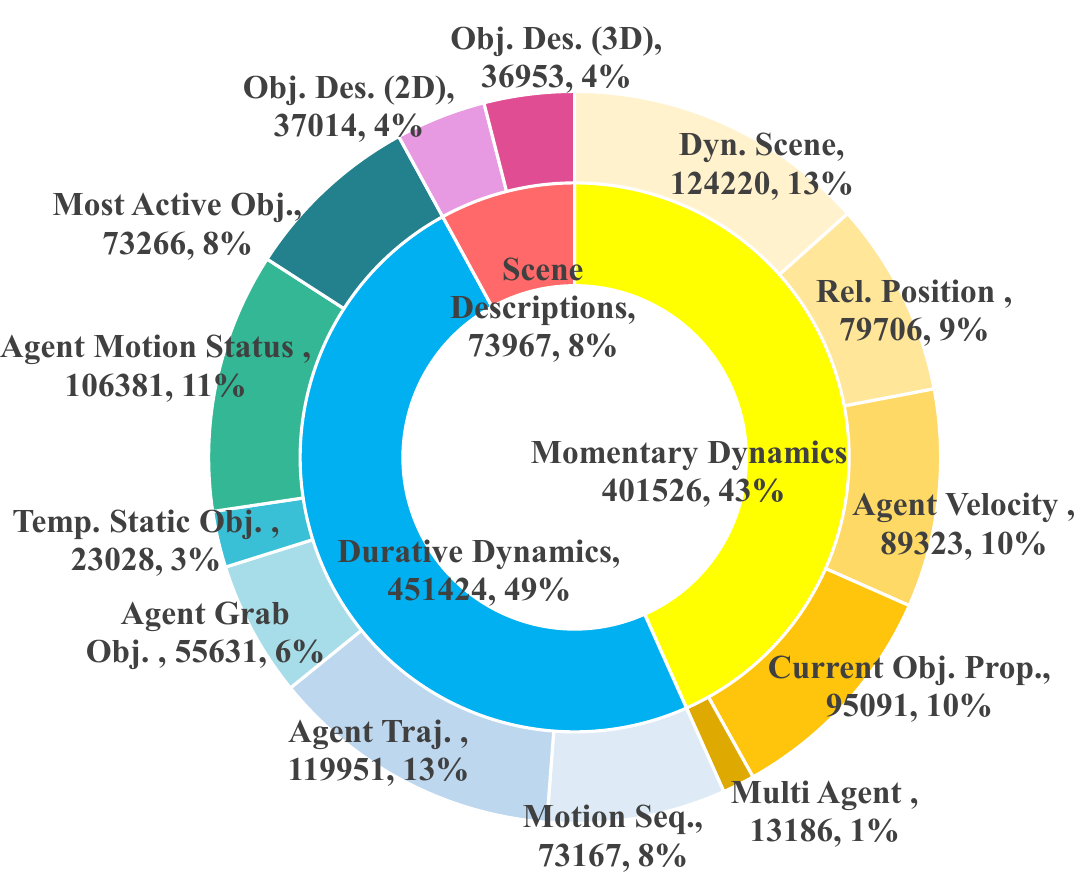}
        \caption{QA pairs in different tasks. We define 12 tasks in 3 domains.}
        \label{fig:category_distribution}
    \end{subfigure}
    \caption{Distribution of data.}
\end{figure*}

Understanding dynamic 4D scenes from an egocentric perspective—modeling changes in 3D spatial structure over time—is crucial for human–machine interaction, autonomous navigation, and embodied intelligence. While existing egocentric datasets contain dynamic scenes, they lack unified 4D annotations and task-driven evaluation protocols for fine-grained spatio-temporal reasoning, especially on motion of objects and human, together with their interactions.
To address this gap, we introduce \textbf{EgoDynamic4D}, a novel QA benchmark on highly dynamic scenes, comprising RGB-D video, camera poses, globally unique instance masks, and 4D bounding boxes. We construct \textbf{927K} QA pairs accompanied by explicit Chain-of-Thought (CoT), enabling verifiable, step-by-step spatio-temporal reasoning. We design 12 dynamic QA tasks covering agent motion, human–object interaction, trajectory prediction, relation understanding, and temporal–causal reasoning, with fine-grained, multidimensional metrics.
To tackle these tasks, we propose an end-to-end spatio-temporal reasoning framework that unifies dynamic and static scene information, using instance-aware feature encoding, time and camera encoding, and spatially adaptive down-sampling to compress large 4D scenes into token sequences manageable by LLMs. Experiments on \textbf{EgoDynamic4D} show that our method consistently outperforms baselines, validating the effectiveness of multimodal temporal modeling for egocentric dynamic scene understanding.

\end{abstract}

\section{Introduction}

With the rapid advancement of embodied AI and human–machine interaction technologies, understanding dynamic 4D scenes from an egocentric viewpoint—integrating 3D spatial dimensions and the temporal dimension—has emerged as a pivotal research challenge. Unlike conventional third-person video analysis~\cite{pang2021pge,song2021co_grounding,pan2021videomoco,li2022from}, egocentric videos exhibit high dynamics, frequent scene changes, and rich interactive behaviors, requiring models not only to capture the wearer’s movement but also to perceive and reason about surrounding people, objects, and their evolving relationships~\cite{ego_exo,Yu2023first_and_third,fan2017first_person,cast,lin2022egocentric,pramanick2023egovlpv2}. Applications such as robotic perception, AR, and autonomous driving demand efficient and accurate egocentric scene understanding to empower the next generation of embodied agents.

\begin{figure*}[ht]
\centering
  \includegraphics[width=1.0\linewidth]{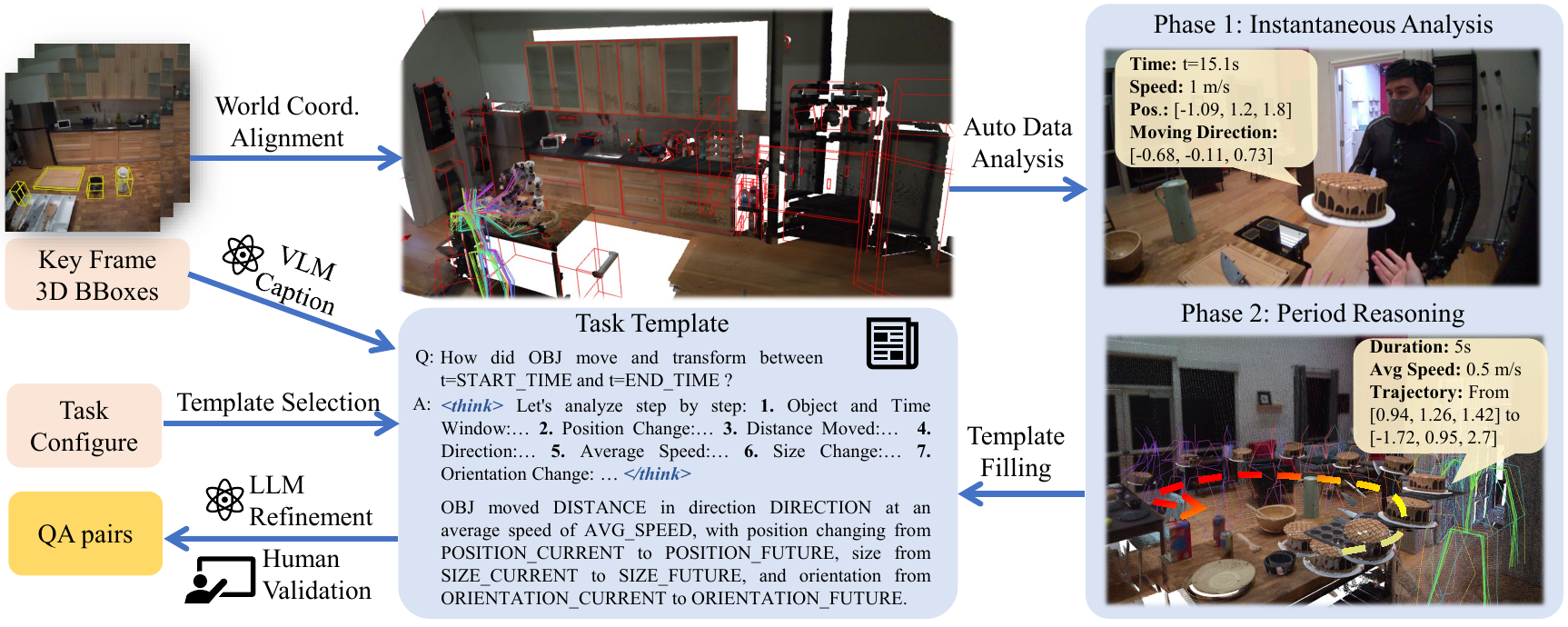}
\caption{QA generation pipeline. Given RGB-D sequences with aligned 3D bounding boxes and poses, we extract spatial-temporal properties, apply template-based CoT reasoning, and refine questions via LLMs and human validation.}
\label{fig:pipeline}
\end{figure*}

Although landmark egocentric datasets Ego4D~\cite{GraumanEgo4D22}, EgoExo4D~\cite{grauman2023egoexo4d}, HD-Epic~\cite{perrett2025hdepic}, and 3D datasets~\cite{dai2017scannet,embodiedscan,chen2020scanrefer} have driven progress in action recognition, object locating, and 3D scene analysis, they suffer from the following key limitations:
(1) Incomplete 4D annotations and lack of dynamic content:  
Egocentric datasets often lack temporally aligned 3D bounding boxes and trajectories, while 3D datasets focus on static scenes without moving objects or agents—limiting the study of real-world dynamics.
(2) Limited evaluation of temporal reasoning:  
Existing benchmarks emphasize short-term or moment-based tasks and lack protocols to assess reasoning over continuous object motion or interaction.
(3) Incomplete multimodal evaluation: 
While some works explore temporal scene graphs~\cite{yang2023psg4d}, they focus on representation construction rather than end-to-end multimodal reasoning, and do not support QA-based evaluation of dynamic 4D scenes.

We propose \textbf{EgoDynamic4D}, the first egocentric QA benchmark for highly dynamic 4D scene understanding. By refining ADT~\cite{pan2023aria} and THUD++~\cite{2024ICRA} annotations, we create a unified, multimodal dataset with RGB-D video, camera poses, globally unique instance masks, and 4D bounding boxes across indoor scenes. It includes 12 QA tasks covering scene descriptions, momentary and durative dynamics, with rich evaluation metrics.
To improve reasoning and transparency, we equip QA pairs with explicit Chain-of-Thought (CoT), enabling step-by-step spatio-temporal explanations that aid training and allow interpretable intermediate results.

We also design an end-to-end spatio-temporal reasoning model that compresses long 4D sequences into LLM-compatible tokens via instance-aware encoding and adaptive down-sampling. Experiments show our method with CoT supervision outperforms prior baselines, setting a new standard for egocentric dynamic scene understanding.

Our main contributions are:
\begin{enumerate}
\item We introduce \textbf{EgoDynamic4D}, a novel highly dynamic 4D scene QA benchmark with 927K QA pairs and explicit CoT, covering 12 task types and multimodal data.
\item We propose an end-to-end spatiotemporal reasoning framework using instance-aware feature encoding, time and camera encoding, and adaptive down-sampling to efficiently handle large 4D data for LLMs.
\item We conduct extensive experiments on EgoDynamic4D using representative models. Our method significantly improves performance and provides interpretable reasoning, establishing a strong baseline for future research.

\end{enumerate}

\section{Related Work}

\paragraph{Egocentric Video Datasets} Egocentric datasets have catalyzed advances in action recognition and object tracking. EpicKitchens~\cite{Damen2018EPICKITCHENS} provides first-person kitchen activity videos but lacks 3D spatial annotations. HD-Epic~\cite{perrett2025hdepic} adds some object localization and short-term action data, but offers only sparse MPS point clouds without dense depth videos, limiting 4D reconstruction. Large-scale datasets like Ego4D~\cite{GraumanEgo4D22}, EgoExo4D~\cite{grauman2023egoexo4d} and EgoLife\cite{yang2025egolife} omit 3D bounding boxes per frame, which constrains dynamic object reasoning. ~\cite{linghu2024multi} focuses on contextual reasoning in 3D scenes rather than dynamic objects. In contrast, ADT~\cite{pan2023aria} and THUD++~\cite{2024ICRA} annotate each frame with RGB-D data, object poses, 2D/3D bounding boxes, and camera trajectories, supporting dynamic scene understanding. EgoDynamic4D builds reorganizing these 2 datasets to create a unified benchmark for dynamic 4D QA tasks.

\paragraph{3DLLMs} 3DLLM~\cite{3dllm} and 3UR-LLM~\cite{xiong20253ur} take point clouds as inputs for tasks like description and localization. LLaVA3D~\cite{zhu2024llava} and Video3DLLM~\cite{zheng2024video3dllm} introduce 3D patches by embedding 2D CLIP features~\cite{radford2021clip} with 3D positional encoding. GPT4Scene~\cite{GPT4Scene} uses BEV images and consistent object IDs to reason about spatial relations. LSceneLLM~\cite{zhi2025lscenellm} employs LLM attention to identify relevant regions for fine details. Chat-3D~\cite{wang2023chat} and Chat-Scene~\cite{huang2024chat} use object identifiers to enable interactive understanding. While effective on static 3D scenes, these approaches lack mechanisms to capture temporal dynamics for 4D reasoning.

\paragraph{4D Scene Understanding Models} Dynamic 4D scene comprehension requires integrating RGB-D information and facilitating cross-time reasoning. PSG4D~\cite{yang2023psg4d} offers a structured 4D scene representation, ~\cite{wu2025learning} leverages 2D annotations to enhance 4D learning via a 3D mask decoder–based LLM. Both focus on graph generation without direct support for QA tasks, whereas our work proposes a unified framework that introduces 4D dynamic scene QA for LLMs.

\paragraph{Spatio‑Temporal CoT}
CoT has been extended to reasoning in vision, like~\cite{yuan2023joint} and ~\cite{yuan2024few}. Video‑CoT~\cite{wang2024videocot} introduces QA pairs with CoT to benchmark reasoning over temporal video content. SpatialCoT~\cite{liu2025spatialcot} targets embodied spatial reasoning by aligning vision–language inputs with spatial coordinates.~\cite{zeng2025futuresightdrive} extends this idea to autonomous driving. Likewise, EgoDynamic4D provides rich CoT data for spatio‑temporal reasoning.

\begin{table*}[ht]
  \centering
  \small
  \begin{tabular}{lccccccc}
    \toprule
    Dataset & Annot. level & Scale & RGB-D & 4D Boxes & Ego. & Dyn. & CoT \\
    \midrule
    HOT3D~\citep{hot3d2024} & Partial 3D & 833 minutes video & Partial & Partial & \Checkmark & \Checkmark & \XSolidBrush\\
    SQA3D~\citep{ma2023sqa3d} & Static 3D & 33.4K questions & Partial & \XSolidBrush& Partial & \XSolidBrush& \XSolidBrush\\
    ScanQA~\citep{Daichi2022ScanQA} & Static 3D & 41K QA pairs & \XSolidBrush& \XSolidBrush& \XSolidBrush& \XSolidBrush& \XSolidBrush\\
    EQA~\citep{yu2019mteqa} & Sparse & 5K questions & \XSolidBrush& \XSolidBrush& \Checkmark & Partial & \XSolidBrush\\
    EmbodiedScan~\citep{embodiedscan} & Static 3D & 1M prompts & \Checkmark & \XSolidBrush& \Checkmark & Partial & \XSolidBrush\\
    HD-EPIC~\citep{perrett2025hdepic} & Partial 3D & 26K questions & \XSolidBrush& Partial & \Checkmark & \Checkmark & \XSolidBrush\\
    ADL4D~\citep{zakour2024adl4d} & Partial 3D & 1.1M frames & \Checkmark & Partial & \Checkmark & \Checkmark & \XSolidBrush\\
    Ego4D~\citep{GraumanEgo4D22} & Sparse & 3,670h, 5K MCQs & \XSolidBrush& \XSolidBrush& \Checkmark & \Checkmark & \XSolidBrush\\
    EgoVQA~\citep{fan2019egovqa} & None & 600 pairs & \XSolidBrush& \XSolidBrush& \Checkmark & \Checkmark & \XSolidBrush\\
    EgoTextVQA~\citep{zhou2025egotextvqa} & None & 7K questions & \XSolidBrush& \XSolidBrush& \Checkmark & \Checkmark & \XSolidBrush\\
    EgoExo4D~\citep{grauman2023egoexo4d} & Sparse & 1,286h & Partial & \XSolidBrush& \Checkmark & \Checkmark & \XSolidBrush\\
    \textbf{EgoDynamic4D(Ours)} & Full 4D & 927K QA pairs & \Checkmark & \Checkmark & \Checkmark & \Checkmark & \Checkmark \\
    \bottomrule
  \end{tabular}
    \caption{Comparison of EgoDynamic4D with spatio-temporal datasets.}
      \label{tab:data_table}
\end{table*}

\section{EgoDynamic4D Benchmark}

\subsection{Overview}

The \textbf{EgoDynamic4D} benchmark in Figure~\ref{fig:dataset-overview} interprets complex scene dynamics, track moving objects, and analyze interactions from a first-person perspective. Given an input 4D egocentric scene \( S = \{F_t\}_{t=1}^T \), where \( F_t = (I_t, D_t) \) represents a frame at timestamp \( t \) with image \( I_t \), depth map \( D_t \), and a natural language question \( Q \) about the scene, the model predicts answer \( A \). The answer \( A \) can be a categorical label, numerical value, or descriptive text, depending on the specific type of question. We organize the QA tasks into three domains:
\textbf{Scene Descriptions} capture semantics of the environment. This includes \textit{object-captioning}, which requires understanding of objects and their surroundings given 2D/3D bounding boxes at specific time.
\textbf{Momentary Dynamics} focus on real-time spatial relations and short-term motion cues. This domain includes object-centric tasks such as \textit{dynamic-scene}, \textit{relative-position}, and \textit{current-object-property}, as well as agent-centric tasks such as \textit{agent-velocity} and \textit{multi-agent-relation}.
\textbf{Durative Dynamics} address longer-term changes and interactions across time. Object-centric tasks include \textit{temporary-static-objects}, \textit{most-active-object}, and \textit{motion-sequence}, while agent-centric tasks include \textit{agent-trajectory}, \textit{agent-grab-object}, and \textit{agent-motion-status}. Detailed definitions are shown in the appendix.

\begin{figure}[!t]
\centering
\includegraphics[width=1\linewidth]{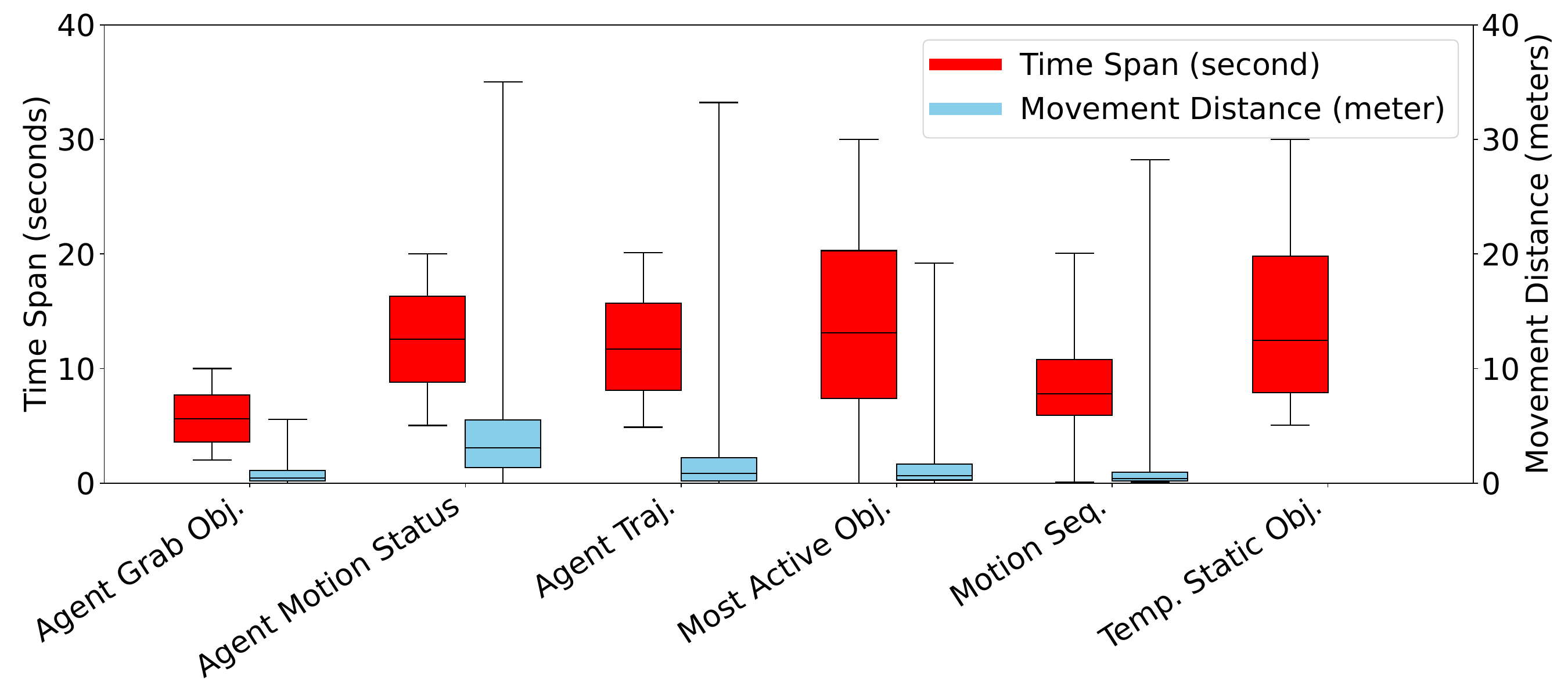}
\caption{Time span and motion distance distribution.}
\label{fig:spatial_span_distribution}
\end{figure}

\subsection{QA Data Construction}
The \textbf{EgoDynamic4D} dataset integrates and enriches two egocentric datasets: ADT (real-world)\cite{pan2023aria} and THUD++ (synthetic scenes)\cite{2024ICRA}, comprising a total of 275 carefully selected sequences, as shown in Figure~\ref{fig:scene_distribution}.
While the overall number of sequences is modest compared to large-scale vision datasets, each sequence was manually selected to ensure rich dynamics and diversity, and is densely annotated with per-frame 3D bounding boxes and fine-grained QA supervision.
The ADT subset contains 236 sequences recorded in real apartment and room environments, covering over 300 static and dynamic objects with high-quality per-frame 3D annotations. The THUD++ subset consists of 39 sequences from synthetic environments (e.g., houses, gyms, offices, supermarkets), with over 100 annotated objects.
As shown in Figure~\ref{fig:temporal_distribution}, the dataset spans a wide range of durations: shorter clips (10–30s) capture momentary dynamics, while longer ones (up to 150s) enable reasoning over extended temporal contexts.

The EgoDynamic4D dataset is constructed through a multi-stage pipeline (Figure~\ref{fig:pipeline}). We first extract synchronized RGB-D frames, 6-DoF camera poses, and aligned 3D bounding boxes for all objects, registering them into a shared world coordinate system for consistent spatio-temporal grounding. Then, we process different domains as follows:
\paragraph{Scene Description} Given 2D projections of 3D bounding boxes on reference frames, we condition Qwen2.5-VL~\cite{bai2025qwen2} on cropped RGB inputs and depth-aware spatial context to generate object-centric and scene-level captions.
\paragraph{Momentary and Durative Dynamics} For dynamic reasoning tasks, we adopt a two-phase QA generation process:
\begin{itemize}
    \item \textit{Frame-level Analysis:} Key frames and adjacent timestamps are analyzed to compute instantaneous properties (e.g., position, velocity, distance, direction) for each object. Using 5 curated templates, we produce QA pairs that query spatial relations and short-term changes.
    \item \textit{Temporal Reasoning:} Longer temporal windows are processed using sliding-window analysis to capture object trajectories, agent-object interactions, and causal dynamics. This stage uses 6 additional templates tailored for questions involving temporal comparisons, forecasting, and interaction reasoning.
\end{itemize}
Each generated QA pair is refined by LLM to improve fluency, consistency, and diversity. We further perform human-in-the-loop verification on the entire corpus to ensure factual correctness and linguistic quality.

\subsection{Data Distribution}

The EgoDynamic4D dataset is distinguished by its diverse and balanced coverage of different kinds of QA tasks and spatial-temporal spans, reflecting the dynamic complexity of egocentric 4D scenes.

\paragraph{Category Distribution}
As shown in Figure~\ref{fig:category_distribution}, the dataset maintains a balanced distribution across 12 task categories, covering both short-term and long-term dynamics. This design enables comprehensive multimodal reasoning from a first-person perspective, with emphasis on object-centric, agent-centric, and agent-object interactions understanding.

\begin{figure*}[ht]
\centering
  \includegraphics[width=1\linewidth]{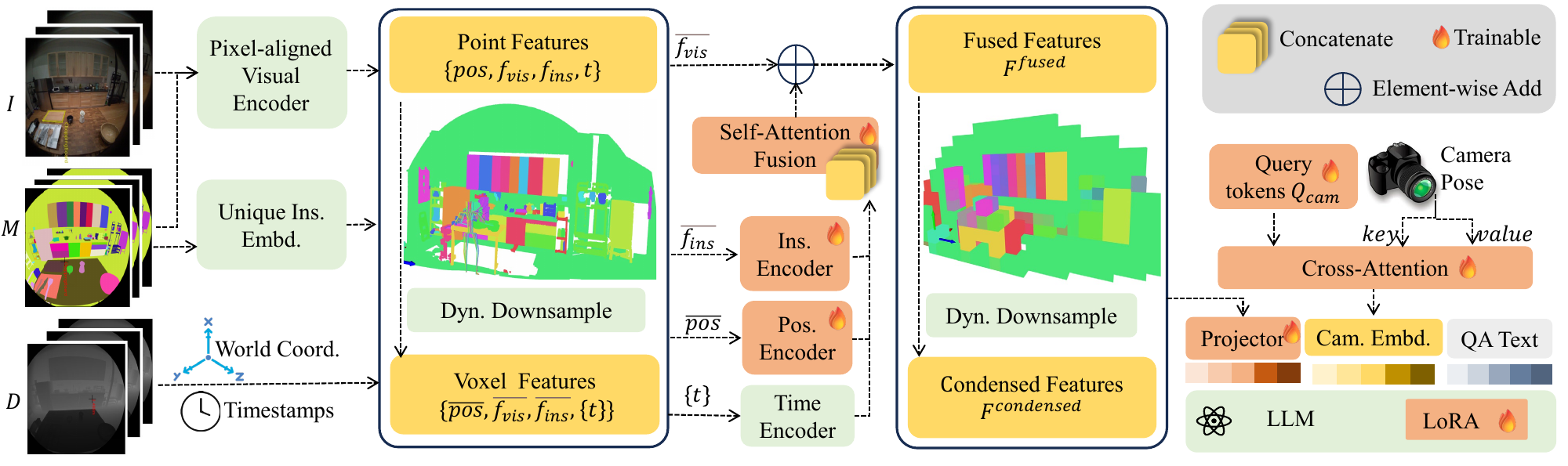}
\caption{The end-to-end framework which encode dynamic 4D scenes based on the egocentric videos.}
\label{fig:method}
\end{figure*}

\paragraph{Spatio-temporal Distribution}
Figure~\ref{fig:spatial_span_distribution} illustrates the distribution of motion distances (blue) and temporal spans (red) across five representative tasks. Most object motions occur within 0--10 meters, while the \textit{most-active-object} task reaches up to 35 meters. Temporally, most events span 10--20 seconds, with longer durations (up to 30 seconds) supporting complex temporal reasoning.

\subsection{EgoDynamic4D vs Prior Datasets}

EgoDynamic4D contains a novel benchmark of 927K QA pairs designed for 12 challenging tasks. Table~\ref{tab:data_table} compares EgoDynamic4D with representative spatio-temporal datasets across key dimensions. Annotation Level (Annot. Level) reflects the granularity of object tracking: Full 4D denotes per-frame 3D bounding boxes over time; other levels (e.g., Partial 3D, Static 3D, Sparse, None) indicate decreasing spatial-temporal completeness. 
Across all frames in EgoDynamic4D, approximately 31.3\% contain at least one dynamic object, with an average of 0.57 dynamic objects and agent interactions present per second. This high temporal density of dynamic events ensures rich supervision for modeling complex egocentric 4D scene dynamics.

\section{Methodology}

We design an end-to-end framework to encode egocentric 4D scenes by fusing spatial and temporal information. The pipeline (Figure~\ref{fig:method}) comprises three stages: instances and timestamps enhanced point-level feature extraction, feature fusion, and projection into tokens for LLM inference. During the whole process, we propose 3 novel component: global unique instance embeddings, voxel timestamps encoding and camera embedding. 

\subsection{Instance and Timestamps Enhanced Point-level Feature Extraction}

To capture fine-grained semantics, we extract per-pixel features from all RGB frames and project them into the 4D dynamic point cloud $P$. Considering the $i$-th frame  in a sequence with $T$ frames in total, let $ I^i \in \mathbb{R}^{H \times W \times 3}, \quad M^i \in \mathbb{R}^{H \times W}, \quad D^i \in \mathbb{R}^{H \times W} $ denote the RGB image, the corresponding instance segmentation map and depth map, respectively. 

\paragraph{Pixel-aligned Visual Encoding} Inspired by Concept-Fusion~\cite{li2022blip} and 3DLLM~\cite{3dllm}, we adapt a similair method on our 4D point cloud. We first use a pretrained vision encoder to compute the global feature representation for the frame:
\begin{equation}
\label{eq: visual-feat}
 F^i_{global} = \mathrm{Pool}(\mathrm{Enc}_{vis}(I^i))  \in \mathbb{R}^{d_{vis}} ,
\end{equation}
where $d_{vis}$ is the dimension of the vision encoder.
For segmentations of $N$ instances in $M^i$, we clip corresponding regions $I^i_j \in \mathbb{R}^{h \times w}\quad (j\in{\{1,2,...,N\}}, h<H, w<W)$, then compute local features similar to Eq~\ref{eq: visual-feat}, getting $ F^i_j \in \mathbb{R}^{d_{vis}} $.
We get the point cloud of the $i$-th frame in world coordinate space $P^i$ . Then, for every single point in the local region $P^i_j$ corresponding to the $j$-th instance segment, we obtain the vision feature $f_{vis}$ by weighted average:
\begin{equation}
\begin{split}
f_{vis} = sim^i_j \cdot F^i_{global}+ (1-sim^i_j) \cdot F^i_j , \\ \text{ 
 where  } sim^i_j = \frac{F^i_j \cdot F^i_{global}}{\|F^i_j\| \cdot \|F^i_{global}\|}.
 \end{split}
\end{equation}

\paragraph{Unique Instance Embedding}We assign each instance a unique embedding vector sampled from $\mathcal{N}(0, I)$, and propagate it across all corresponding points in the segmentation mask. Let $d_{ins}$ be the embedding dimension, then each point’s instance encoding is $f_{ins} \in \mathbb{R}^{d_{ins}}$. In high-dimensional spaces, randomly sampled vectors are nearly orthogonal with high probability~\cite{ghojogh2021johnson}, enabling a large number of identities to be distinguished. This enriches point features with instance identity information. 

\begin{table*}[ht]
  \small
  \setlength\tabcolsep{1.2pt} %
  \centering
  \begin{tabular}{ll|c|cc ccc|cc cccc|c}
    \toprule
\multirow{2}{*}[-3ex]{Subset} & \multirow{2}{*}[-3ex]{Method} & \multicolumn{1}{c}{Scene Des.} & \multicolumn{5}{c}{Momentary Dynamics} & \multicolumn{5}{c}{Durative Dynamics} & & \multirow{2}{*}[-3ex]{\textbf{\makecell[c]{Overall\\BLEU-4}}} \\
   & & \makecell{obj.\\caption\\(bleu4)\textuparrow} & 
   \makecell{dyn.\\scene\\(f1)\textuparrow} & \makecell{rel.\\pos.\\(acc\%)\textuparrow} & \makecell{curr.\\obj.\\prop.\\(acc\%)\textuparrow} & \makecell{agent\\vel.\\(acc\%)\textuparrow} & \makecell{multi\\agent\\rel.\\(acc\%)\textuparrow} & 
   \makecell{motion\\seq.\\(acc\%)\textuparrow} & \makecell{most\\active\\obj.\\(acc\%)\textuparrow} & \makecell{temp.\\static\\obj.\\(f1)\textuparrow} & 
    \makecell{agent\\traj.\\(acc\%)\textuparrow} & \makecell{agent\\motion\\status\\(acc\%)\textuparrow} & \makecell{agent\\grab\\obj.\\(acc\%)\textuparrow} & \\
    \midrule
 \multirow{9}{*}{ADT} 
 & LLaVA-3D & 0.072 & 0.290 & 42.56 & 30.12 & 23.07 & 29.94 & 25.78 & 28.62 & 0.718 & 24.21 & 24.30 & 12.30 & 0.388 \\
 & Video3DLLM & 0.034 & 0.307 & 35.65 & 27.97 & 24.55 & 22.31 & 23.80 & 27.47 & \underline{0.737} & 24.07 & 23.22 & 8.34 & 0.392 \\
 & VG-LLM & 0.211 & 0.297 & 43.54 & 43.72 & \underline{25.95} & 29.52 & 26.48 & 29.58 & 0.713 & 26.51 & 26.02 & 16.38 & 0.406 \\
 & 3DLLM & 0.033 & 0.003 & 30.48 & 14.52 & 20.49 & 14.78 & 17.69 & 2.35 & 0.502 & 6.96 & 18.18 & 5.96 & 0.345 \\
 & LL3DA & 0.014 & 0.006 & 14.68 & 0.90 & 23.21 & 17.04 & 18.78 & 8.38 & 0.166 & 18.07 & 21.29 & 1.49 & 0.287 \\
 & Chat-Scene & 0.000 & 0.104 & 39.60 & 0.00 & 8.25 & 29.39 & 0.00 & 23.28 & 0.628 & 8.13 & 0.00 & 8.98 & 0.187 \\
 & \textbf{Ours} & \textbf{0.244} & \textbf{0.455} & \underline{49.79} & \underline{58.39} & \textbf{31.32} & \underline{55.47} & \underline{40.56} & \underline{41.98} & \textbf{0.763} & \underline{46.11} & \textbf{29.94} & \underline{28.25} & \underline{0.435} \\
 & \textbf{Ours+CoT} & \underline{0.238} & \underline{0.454} & \textbf{84.11} & \textbf{67.35} & 19.33 & \textbf{65.72} & \textbf{57.04} & \textbf{56.82} & 0.726 & \textbf{47.35} & \underline{27.69} & \textbf{29.64} & \textbf{0.436} \\
\midrule
 \multirow{9}{*}{THUD++} 
 & LLaVA-3D & 0.034 & 0.152 & 10.99 & 9.46 & 45.91 & -- & 11.01 & 10.03 & 0.436 & 0.80 & 37.60 & -- & 0.370 \\
 & Video3DLLM & 0.040 & 0.087 & 9.66 & 3.11 & 44.46 & -- & 9.47 & 3.37 & 0.410 & 1.63 & 36.81 & -- & 0.349 \\
 & VG-LLM & 0.083 & 0.002 & 9.41 & 1.55 & 44.86 & -- & 10.26 & 4.07 & 0.014 & 1.41 & 39.85 & -- & 0.354 \\
 & 3DLLM & 0.009 & 0.000 & 9.82 & 0.00 & 36.69 & -- & 10.88 & 0.69 & 0.000 & 3.57 & 31.87 & -- & 0.312 \\
 & LL3DA & 0.002 & 0.000 & 6.66 & 0.00 & 32.91 & -- & 7.08 & 0.60 & 0.000 & 1.90 & 24.89 & -- & 0.265 \\
 & Chat-Scene & 0.000 & 0.024 &  \underline{13.70} & 0.00 & 20.55 & -- & 0.00 & 19.12 & 0.386 & 3.42 & 0.00 & -- & 0.185 \\
 & \textbf{Ours} & \underline{0.097} & \underline{0.221} & 13.14 & \underline{27.68} & \textbf{59.24} & -- & \underline{26.10} &  \underline{21.89} & \underline{0.472} & \underline{7.36} &  \underline{50.42} & -- & \underline{0.403} \\
 & \textbf{Ours+CoT} & \textbf{0.105} & \textbf{0.333} & \textbf{61.66} & \textbf{65.49} &  \underline{48.59} & -- & \textbf{43.67} & \textbf{32.66} & \textbf{0.596} & \textbf{18.33} & \textbf{55.58} & -- & \textbf{0.431} \\
   \bottomrule
  \end{tabular}
\caption{Baseline models, best results in \textbf{bold} and the second with \underline{underline}.}
\label{tab:baseline}
\end{table*}

\paragraph{Timestamps} Each point is marked with its timestamp $t$. Combining position $pos \in \mathbb{R}^3$ with visual features, instance embeddings, and timestamps, the final representation of each point $p$ is $f_p = {\{ pos_p,f_{vis,p},f_{ins,p},t_p\}}$. Since points within a segmented region share identical features per frame, the enhanced local point cloud representation is $S^i_j = {\{f_p \mid p \in P^i_j\}} \in \mathbb{R}^{|P|\times (3+d_{vis}+d_{ins}+1)}$, where $|P|$ denotes the number of points. 
Thus, the full point cloud representation $S$ integrates 4D information.

\subsection{Feature Fusion}
To reduce the computational complexity of the sparse feature $S$, we apply an octree-based dynamic downsampling on spatial positions, and voxel features are aggregated per node. For each node, point positions, visual features, and instance embeddings are averaged, while all timestamps are collected into a set $\{t\}$. This yields voxels $V$, each represented by $f_v = \{\overline{pos_v}, \overline{f_{vis,v}}, \overline{f_{ins,v}}, {\{t\}}_v\}, v \in V$. The process reduces voxel count from 50M–300M to 100K–250K.

\paragraph{Time Encoding} We encode timestamps to capture dynamic evolution. Let $ \{t_{v,1}, ...,t_{v,k},..., t_{v,K}\} $ be the $K$ times when points appears in the voxel $v$. 
Then compute sinusoidal features $s_{v,k}$ of $t_{v,k}$ as Eq~\ref{eq:time}:  
\begin{equation}
\begin{split}
\label{eq:time}
 m\in {\{0,1,...,\lfloor \tfrac{d_{vis}}{2} \rfloor-1\}},\;
    d_m = \exp\!\bigl(-\tfrac{\ln 10^4}{d_{vis}}\,m\bigr), \\
    s_{v,k}^{2m} = \sin\bigl(t_{v,k}\cdot d_m\bigr), \;
s_{v,k}^{2m+1} = \cos\bigl(t_{v,k}\cdot d_m\bigr).
\end{split}
\end{equation}
Then we aggregate them via max and mean pooling as Eq~\ref{eq:add}, where $\alpha$ is the factor. This Fourier basis supports distinguishing intervals up to $ 10^4 $ s.
\begin{equation}
\label{eq:add}
t_v^{emb} = \alpha \cdot \max\limits_{k} s_{v,k}+(1-\alpha)  \cdot \operatorname*{avg}\limits_{k} s_{v,k} \in \mathbb{R}^{d_{vis}}
\end{equation}

\paragraph{Instance Encoding and Feature Integration} A learnable matrix $ W_{ins}\in\mathbb{R}^{d_{vis}\times d_{ins}} $ projects instance embedding $\overline{f_{vis,v}}$ into hidden dimension, then it is fused with time embeddings and original position embeddings via self-attention (SA), adding to visual feature together.
 \begin{equation}
 \begin{split}
 \label{eq:fuse}
f_v^{fused} =  \overline{f_{vis,v}} + \\ \mathrm{SA}\bigl([\,W&_{ins}\cdot\overline{f_{ins,v}} \, \| \, t_v^{emb} \| \mathrm{Enc}_{pos}(\overline{pos_v}) \, ]\bigr)
 \end{split}
\end{equation}

After obtaining the fused voxel features $F^{fused} \in \mathbb{R}^{|V|\times{d_{vis}}}$, where $|V|$ is the number of voxels, we apply another downsampling to yield condensed representation $F^{condensed}$ ($\sim$1K tokens) for LLM processing.

\begin{table*}[ht]
\small
\setlength\tabcolsep{1.5pt} %
\centering
\begin{tabular}{ll|c|cc ccc|cc cccc|c}
\toprule
\multirow{2}{*}[-3ex]{Subset} & \multirow{2}{*}[-3ex]{Settings} & \multicolumn{1}{c}{Scene Des.} & \multicolumn{5}{c}{Momentary Dynamics} & \multicolumn{5}{c}{Durative Dynamics} & & \multirow{2}{*}[-3ex]{\textbf{\makecell[c]{Overall\\BLEU-4}}} \\
& & \makecell{obj.\\caption\\(bleu4)\textuparrow} & 
\makecell{dyn.\\scene\\(f1)\textuparrow} & \makecell{rel.\\pos.\\(acc\%)\textuparrow} & \makecell{curr.\\obj.\\prop.\\(acc\%)\textuparrow} & \makecell{agent\\vel.\\(acc\%)\textuparrow} & \makecell{multi\\agent\\rel.\\(acc\%)\textuparrow} & 
\makecell{motion\\seq.\\(acc\%)\textuparrow} & \makecell{most\\active\\obj.\\(acc\%)\textuparrow} & \makecell{temp.\\static\\obj.\\(f1)\textuparrow} & 
\makecell{agent\\traj.\\(acc\%)\textuparrow} & \makecell{agent\\motion\\status\\(acc\%)\textuparrow} & \makecell{agent\\grab\\obj.\\(acc\%)\textuparrow} & \\
\midrule
\multirow{5}{*}{ADT} & whole & $\textbf{0.244}$ & $\textbf{0.454}$ & $\textbf{49.79}$ & $\textbf{58.39}$ & $\textbf{31.32}$ & $\textbf{55.47}$ & $\textbf{40.56}$ & $\textbf{41.98}$ & $\textbf{0.763}$ & $\textbf{46.11}$ & $\textbf{29.94}$ & $\textbf{28.25}$ & $\textbf{0.435}$ \\
& w/o c & $\underline{0.239}$ & $\underline{0.450}$ & $\underline{49.16}$ & $\underline{48.72}$ & $30.34$ & $\underline{54.11}$ & $\underline{39.52}$ & $40.74$ & $0.757$ & $\underline{43.82}$ & $28.58$ & $\underline{24.70}$ & $\underline{0.432}$ \\
& w/o c\&i & $0.237$ & $0.433$ & $48.50$ & $48.39$ & $30.12$ & $52.49$ & $37.47$ & $39.60$ & $0.757$ & $42.75$ & $28.03$ & $24.59$ & $0.429$ \\
& w/o c\&i\&t & $0.234$ & $0.346$ & $48.37$ & $37.30$ & $26.72$ & $42.41$ & $31.95$ & $33.37$ & $0.739$ & $31.22$ & $25.55$ & $19.34$ & $0.411$ \\
& mlp & $0.236$ & $0.438$ & $48.94$ & $45.92$ & $\underline{30.55}$ & $53.11$ & $36.18$ & $\underline{41.59}$ & $0.754$ & $43.72$ & $\underline{28.73}$ & $23.75$ & $0.429$ \\
\midrule
\multirow{5}{*}{THUD++} & whole & $\textbf{0.097}$ & $\textbf{0.221}$ & $\underline{13.14}$ & $\underline{27.68}$ & $\textbf{59.24}$ & -- & $\textbf{26.10}$ & $\textbf{21.89}$ & $\textbf{0.472}$ & $\textbf{7.36}$ & $\underline{50.42}$ & -- & $\textbf{0.403}$ \\
& w/o c & $\underline{0.095}$ & $\underline{0.201}$ & $\underline{13.14}$ & $26.55$ & $57.67$ & -- & $\underline{21.18}$ & $17.30$ & $0.424$ & $\underline{6.55}$ & $48.67$ & -- & $\underline{0.400}$ \\
& w/o c\&i & $0.087$ & $0.189$ & $12.22$ & $24.58$ & $57.63$ & -- & $18.20$ & $16.96$ & $0.411$ & $5.31$ & $48.45$ & -- & $0.395$ \\
& w/o c\&i\&t & $0.071$ & $0.176$ & $11.91$ & $22.46$ & $55.77$ & -- & $17.11$ & $16.87$ & $0.402$ & $3.90$ & $48.36$ & -- & $0.394$ \\
& mlp & $0.093$ & $0.195$ & $12.99$ & $\textbf{31.07}$ & $\underline{59.10}$ & -- & $20.04$ & $\underline{18.86}$ & $\underline{0.448}$ & $6.28$ & $\textbf{52.09}$ & -- & $0.399$ \\
\bottomrule
\end{tabular}
\caption{Ablation on camera  (c), instance (i), time (t) embedding, and MLP feature fusion (mlp, with c\&i\&t).}
\label{tab:ablation}
\end{table*}

\begin{figure*}[ht]
\centering
\begin{subfigure}[b]{0.48\linewidth}
\centering
\includegraphics[width=\linewidth]{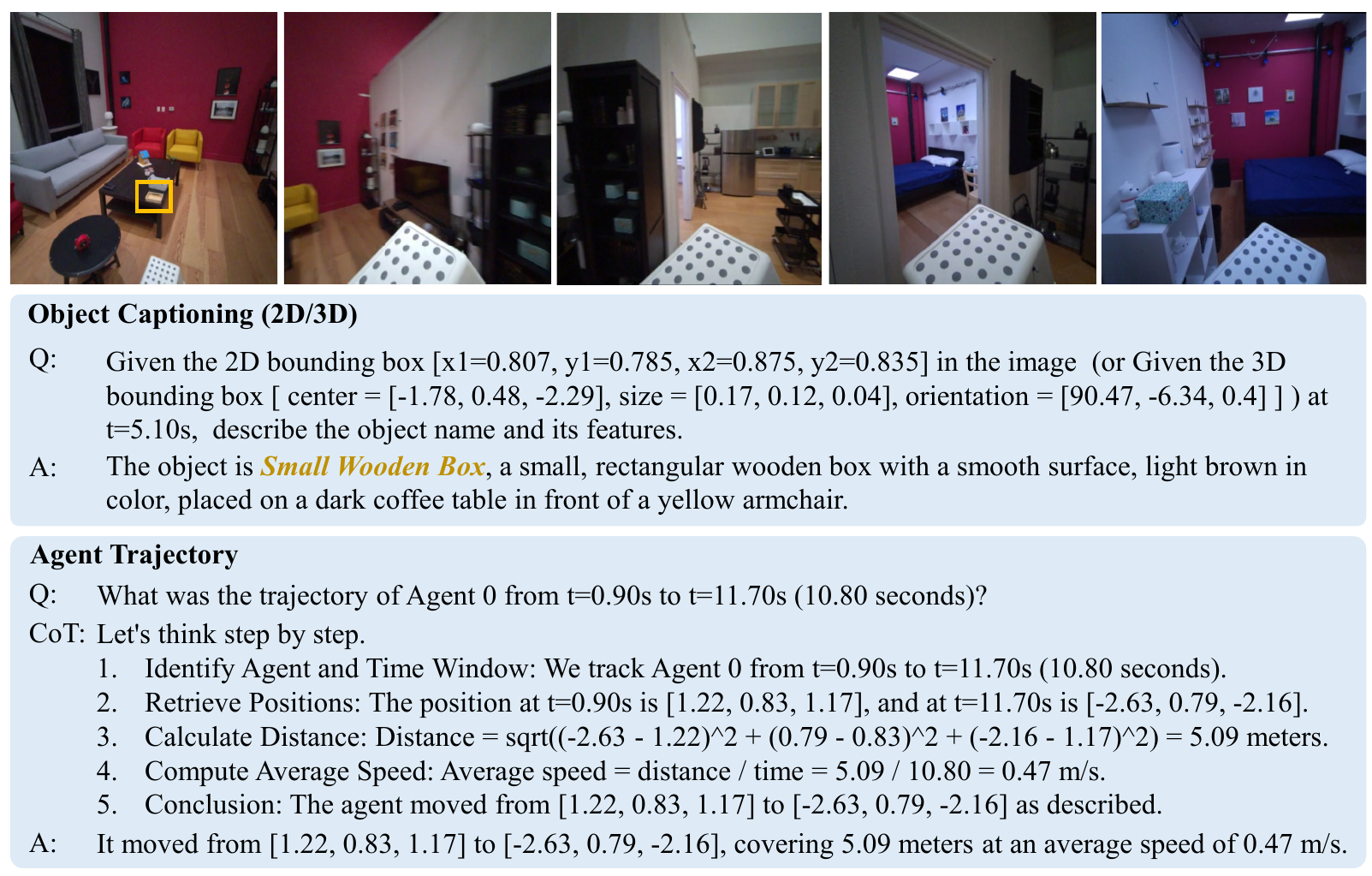}
\caption{Examples of \textit{object-captioning} and \textit{agent-trajectory} tasks.}
\label{fig:example_data}
\end{subfigure}
\hspace{0.01\linewidth}
\begin{subfigure}[b]{0.5\linewidth}
\centering
\includegraphics[width=\linewidth]{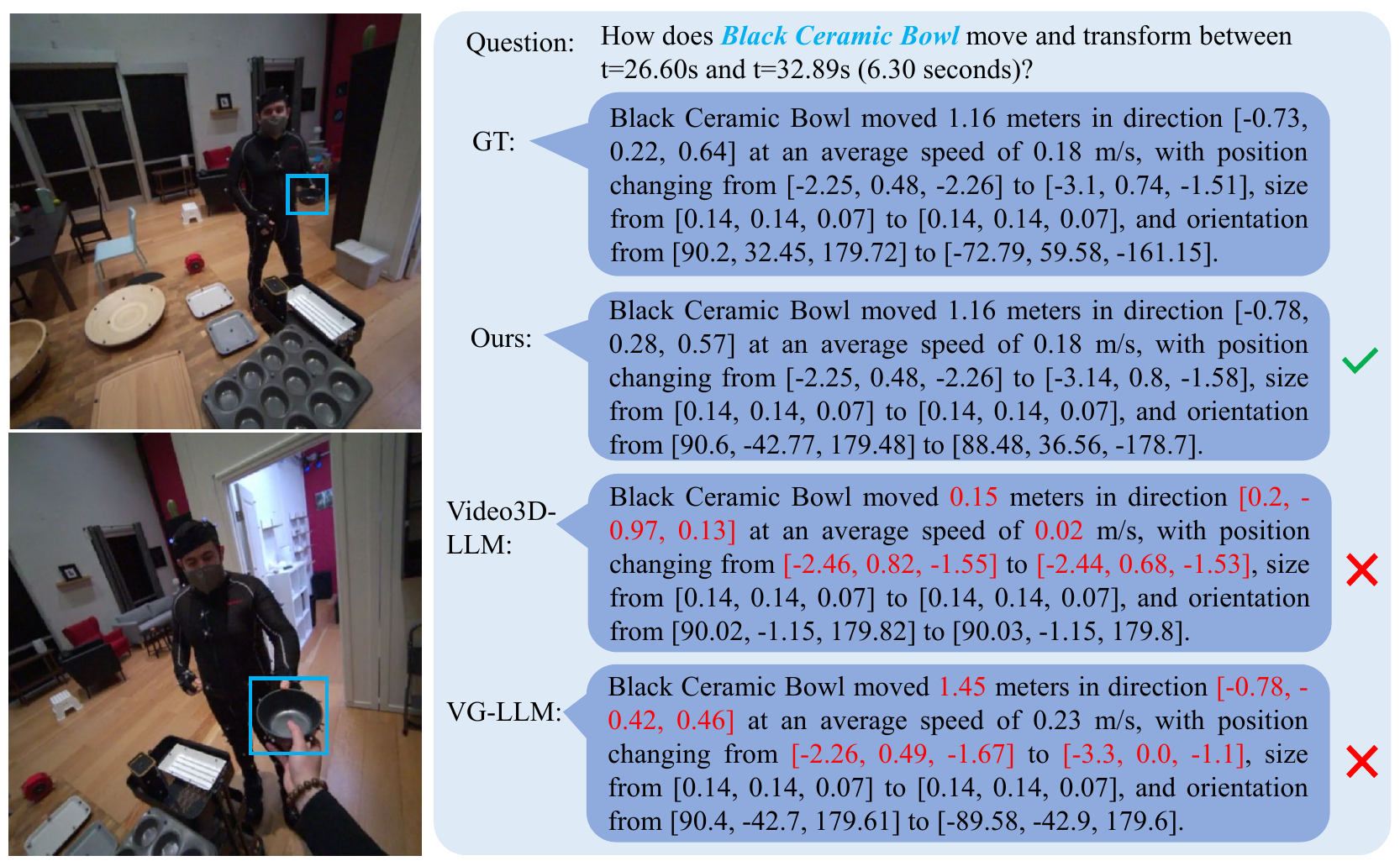}
\caption{Comparison of baselines on \textit{motion-sequence} task.}
\label{fig:example_model}
\end{subfigure}
\caption{Qualitative examples illustrating the complexity of spatial-temporal reasoning in EgoDynamic4D.}
\label{fig:example_qualitative}
\end{figure*}

\paragraph{Camera Embedding}
We compress the sequence of camera poses $ (x, y, z, q_x, q_y, q_z, q_w) \in \mathbb{R}^7 $ into a compact embedding using a learnable attention mechanism. A sequence of $T$ poses are first project into hidden space $f_{cam} \in \mathbb{R}^{T \times  d_{vis}} $, and then $M$ learnable query tokens $ Q_{cam} \in \mathbb{R}^{M \times d_{vis}} $ attends to $f_{cam}$ through cross-attention (CA):
\begin{equation}
F_{cam} = \mathrm{CA}(Q_{cam}, f_{cam}, f_{cam}) \in \mathbb{R}^{M \times d_{vis}}
\end{equation}
Finally, $F^{condensed}$ and $ F_{cam} $ are projected into LLM embedding space, then we use LoRA for efficient fine-tuning.

\section{Experiments}

\subsection{Implementation Details}

We conduct experiments on  EgoDynamic4D (80\% train, 20\% test). Our model builds upon LLaVA3D (CLIP + LLaMA)~\cite{zhu2024llava} with frozen backbones, unfreezing only the proposed modules ($d_{ins}$=8, $M$=8) and LoRA parameters (rank=8, alpha=16). We set sample fps to 5, and train using AdamW (learning rate 5e-5) for 2 epochs on 8 × RTX 4090 (24GB) with batch size 1 per GPU. 

\subsection{Results on EgoDynamic4D}
We evaluate our method on the EgoDynamic4D dataset (ADT and THUD++ subsets), and benchmark it against representative baselines. Since 4D LLMs like LLaVA-4D~\cite{zhou2025llava} are not publicly available yet, we evaluated 3D LLMs including LLaVA-3D~\cite{zhu2024llava}, Video3DLLM~\cite{zheng2024video3dllm}, LL3DA~\cite{chenLL3DA2024}, Chat-Scene~\cite{huang2024chat},  3DLLM~\cite{3dllm} and VG-LLM~\cite{2025vgllm}. To ensure a fair comparison under limited resources, we fine-tune all LLM models with the same LoRA configuration, while keeping other components in original implementation settings. Evaluation follows the benchmark's protocol (details in the appendix), with thresholds settings: speed error within $0.05~m/s$, direction error within $0.5~rad$, IOU above $0.1$, distance and position errors within $0.1~m$.

As shown in Table~\ref{tab:baseline}, our models consistently outperform all baseline methods on EgoDynamic4D  across both subsets, demonstrating robust spatial-temporal reasoning capabilities. Our CoT-enhanced model achieves top performance in overall BLEU-4 and most tasks, showcasing its strength in capturing complex 4D scene dynamics and relationships. Our non-CoT model also surpasses baselines, delivering competitive results across all metrics. This consistent superiority highlights the effectiveness of our approach in handling diverse 4D environments, with CoT further enhancing precision in complex spatial-temporal tasks.

\subsection{Ablation Study}
Table~\ref{tab:ablation} evaluates camera, instance, time embeddings, and also compares attention-based feature fusion (whole) with MLP-based fusion (mlp) on EgoDynamic4D. The full model achieves the highest overall BLEU-4, excelling in dynamic tasks. On THUD++, where motion patterns in \textit{current-object-property} and \textit{agent-motion-status} are less diverse than in ADT, the MLP-based fusion outperforms the full model, likely due to MLP’s localized feature fusion preserving fine-grained details in less dynamic tasks, whereas attention’s global integration may introduce noise. Ablations confirm that camera, instance, and time embeddings enhance spatial, object-specific, and temporal reasoning.

\subsection{Qualitative Analysis}
Figure~\ref{fig:example_data} showcases examples from the dataset, including \textit{object-captioning} given 2D or 3D bounding boxes, and \textit{agent-trajectory} that is solved using CoT.
Figure~\ref{fig:example_model}  compares our model against other baselines on \textit{motion-sequence} task. The results highlight the challenging nature of EgoDynamic4D benchmark and the effectiveness of our method in capturing fine-grained dynamics.

\section{Discussion and Conclusion}
We introduced \textbf{EgoDynamic4D}, a novel benchmark for highly dynamic egocentric 4D scene understanding, featuring 927K QA pairs across 12 diverse task types that capture dense, fine-grained temporal dynamics. To support spatio-temporal reasoning, we proposed a unified framework incorporating egocentric pose embedding, global instance encoding, and Fourier-based timestamp embedding. An octree-based downsampling strategy ensures efficiency while preserving structural integrity. Experiments show strong performance, though challenges remain in generalization and robustness to degraded point clouds. We leave these directions for future work. EgoDynamic4D lays a new foundation for egocentric 4D understanding, with broad relevance to embodied AI and robotics.

\section*{Acknowledgments}

This work was sponsored by the National Key R\&D Program of China (No. 2024YFF0907803), the National Natural Science Foundation of China (No. 62576308), and the Fundamental Research Funds for the Central Universities (No. 226-2025-00167).

\small
\bibliography{ref}

\end{document}